\documentclass[letterpaper, 10 pt, conference]{ieeeconf}  

\IEEEoverridecommandlockouts                              

\usepackage{booktabs}
\usepackage{multirow}
\usepackage{tabularx}  
\usepackage[export]{adjustbox}
\usepackage{graphicx}
\usepackage{amsmath} 
\usepackage{amssymb}  

\usepackage{caption}

\usepackage[bookmarks=true,colorlinks=true,citecolor=magenta,linkcolor=cyan]{hyperref}
\usepackage{hypcap}

\usepackage{booktabs} 
\usepackage{makecell}
\usepackage{pifont}
\usepackage{float}
\usepackage{placeins}
\usepackage{subcaption}
\usepackage{array}
\usepackage{siunitx}

\usepackage[ruled,vlined]{algorithm2e}

\let\labelindent\relax
\usepackage{enumitem}
\usepackage{arydshln}
\usepackage{wrapfig}

\usepackage[nospace]{cite}
\usepackage{ifthen}
\usepackage{xcolor}
\usepackage{multicol}

\definecolor{bccolor}{HTML}{E34A6F}
\definecolor{daggercolor}{HTML}{FFBE00}
\definecolor{rlcolor}{HTML}{2398DA}
\definecolor{longcolor}{HTML}{6B529C}
\definecolor{cotrain}{HTML}{39BC56}

\newcommand{\appref}[1]{\hyperref[#1]{App.~\ref{#1}}}
\newcommand{\refeq}[1]{\hyperref[#1]{Eq.~(\ref{#1})}}

\newboolean{draft-mode}
\setboolean{draft-mode}{false}

\usepackage{balance}

\title{\LARGE \bf
Bridging the Sim2Real Gap:\\ A Framework for Evaluating Vision Encoder Pre-Training for Visuomotor Policy Transfer
}

\author{Yash Yardi$^{*\dagger}$ Samuel Biruduganti$^{*\dagger}$ Lars Ankile$^{\ddagger}$\vspace{5pt}\\ $^\dagger$University of Illinois Urbana-Champaign $^\ddagger$Massachusetts Institute of Technology\\\texttt{\{yyard, sbiru3\}@illinois.edu ankile@mit.edu}\thanks{$^*$Equal Contribution}}

\begin{document}

\let\oldtwocolumn\twocolumn
\renewcommand\twocolumn[1][]{%
    \oldtwocolumn[{#1}{
    \vspace{-20pt}
    \begin{flushleft}
           \centering
    \end{flushleft}
    }]
}

\maketitle
\thispagestyle{empty}
\pagestyle{empty}

\begin{abstract}
Simulation offers a scalable and efficient alternative to real-world data collection for learning visuomotor robotic policies. However, the simulation-to-reality, or ``Sim2Real'' distribution shift---introduced by employing simulation-trained policies in real-world environments---frequently prevents successful policy transfer. We present an offline framework to evaluate the performance of using large-scale pre-trained vision encoders to address the Sim2Real gap. We examine a diverse collection of encoders, assessing their ability to extract features necessary for robot control (Action Score) while remaining invariant to task-irrelevant environmental variations (Domain Invariance Score). Evaluating 23 encoders, we reveal patterns across architectures, pre-training datasets, and parameter scales. Our findings show that manipulation-pretrained encoders consistently achieve higher Action Scores, CNN-based encoders demonstrate stronger domain invariance than ViTs, and the best-performing models combine both properties, underscoring DIS and AS as complementary predictors of Sim2Real transferability.


\end{abstract}


\begin{keywords}
Sim2Real, vision encoder, transfer learning, robotic manipulation, feature extraction, domain invariance
\end{keywords}



\section{Introduction}
Robots that can reliably function in unstructured environments, operating from vision alone, have the potential to transform daily life and industry  \cite{autonrobots}. Visuomotor robotic policies---algorithms enabling robots to perform tasks solely using image observations---are necessary for this. Recent advancements have been made using Neural Networks (NNs) and machine learning to learn these policies from data \cite{chi2024diffusionpolicyvisuomotorpolicy, ssv_visuomotorlearning, realworldrobotlearningmasked}. However, gathering the vast data necessary for robots to learn robust policies on physical robots is often impractical and unsafe.

Simulation is a scalable alternative that allows researchers to rapidly replicate years of robot training \cite{deepreinforcementlearning}. Despite the incredible benefits, a simulated environment will inevitably differ from its real-world counterpart. This difference introduces a distribution shift known as the Sim2Real gap, which, if not adequately addressed, results in poorly performing simulation-trained policies when deployed in real environments.

Prior work has demonstrated that pre-trained vision encoders improve Sim2Real transfer, however, a central challenge remains: how can we evaluate representation quality offline, before expensive robot rollouts \cite{renaissanceinvestigatingpretrainingvisionlanguage}. These neural networks, pre-trained on enormous datasets, convert raw images into compact, information-rich representation vectors that improve transferability by preferably extracting task-relevant information. This study assesses various encoders' potential for transfer learning of visuomotor policies. We hypothesize that effective vision encoders capable of bridging the Sim2Real gap should pick out relevant features from an image to infer actions (e.g., the objects and their position) and ignore task-irrelevant parts of the image that change between simulated and real environments (e.g., the table color and clutter objects).

In this paper, we propose a framework for offline evaluation of encoders in Sim2Real contexts. The framework is built around two quantitative metrics: the Domain Invariance Score (DIS), which measures the alignment of simulated and real-world embeddings, and the Action Score (AS), which evaluates how well encoder embeddings capture task-relevant information through linear probing. We incorporate Grad-CAM saliency maps as a qualitative validation tool, providing interpretable visual evidence that high-performing encoders attend to task-relevant features across both simulated and real domains.  



\section{Related Works}
Given the significant potential of AI-enabled robots capable of real-world unsupervised operation, studies have explored the nature of the Sim2Real gap and devised potential solutions. 


\subsection{Pre-Trained Visual Representations}
The robotics research community has investigated pre-trained visual representations (PVRs) for training downstream policies in real-world tasks, comparing their performance across simulated and real environments \cite{surveypretrainedencoders}. PVRs produce embeddings (output features of 1-dimensional vectors) extracted from visual data using pre-trained vision encoders designed for visuomotor tasks. Silwal et al. \cite{surveypretrainedencoders} conducted a comprehensive investigation into using PVRs for training policies, evaluating their effectiveness in simulated and real environments. The study demonstrated that PVRs significantly improved Sim2Real transfer in the ImageNav task compared to previous methods.

Burns et al. \cite{PVRs} further studied the success of PVRs in robotic manipulation tasks, focusing on the attributes of pre-trained vision models robust to distribution shifts, including subtle lighting and texture changes. The study evaluates fifteen models to identify which features made them more reliable under these conditions. The authors found that models pre-trained on manipulation-relevant data could consistently generalize better than models trained on standard pre-training datasets like ImageNet \cite{5206848}. Notably, they found that a model's strong emergent segmentation ability---the capacity to separate essential parts of an image from irrelevant details naturally---showcased its consistent success in generalization despite distribution shifts. Given the significant relevance of these distribution shifts to the Sim2Real gap, this finding applies directly to this inquiry's application of the PVRs in examining whether the nature of the model's pre-training indicates its performance.

\subsection{Custom Frameworks for Learning to Generalize}
To combat the distribution shift, a study introduced Maniwhere, a framework to improve generalization across diverse visual disturbances \cite{maniwhere}. Maniwhere leverages multi-view representation learning with a Spatial Transformer Network to extract meaningful information from varying viewpoints. By combining a fixed image with a randomly perturbed one, the model learns correspondences that enhance its ability to generalize semantic features. Then, reinforcement learning is employed to train the robot through environmental interaction, optimizing its decision-making process. Evaluated on eight manipulation tasks, the unique framework demonstrated superior visual generalization and Sim2Real transfer compared to other designs. This study's multi-view approach highlights a promising avenue for addressing the Sim2Real gap in future research.


\section{Methods} \label{methods}
We develop a framework to evaluate several pre-trained vision encoders' ability to learn effective visuomotor policies transferable to real-world applications. We posit that to tackle the Sim2Real gap, effective encoders should exhibit high action-inference accuracy---indicating relevant feature extraction ability---and low source determination ability, or high domain invariance, measured through the Domain Invariance Score (DIS). Encoder attributes are assessed by generating embeddings, using our framework of linear probes and distance metrics for quantitative comparison using the Action Score (AS) \cite{linearprobe}. 

With matching action and domain labels, the image dataset utilized more than 50,000 frames of a robot arm trying to construct a small table from simulated and real-world environments \cite{heo2023furniturebenchreproduciblerealworldbenchmark}. This large collection offers a solid basis for assessing encoder generalization across domain shifts, including different illumination and object layouts. 
\cite{ankile2024imitationrefinementresidual}

\begin{figure}[!htbp]
    \centering
    \includegraphics[width=\linewidth]{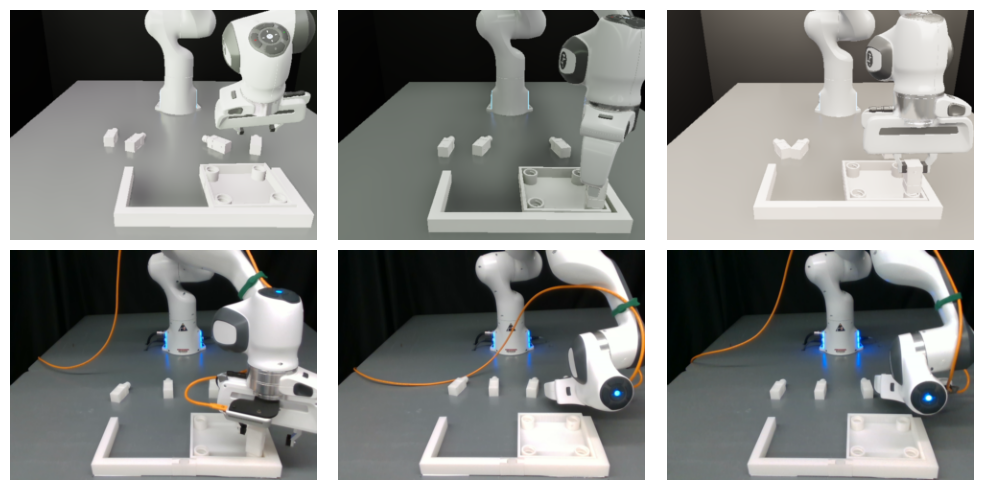}
    \caption{Three Random Frames from the Dataset used. Top Row is Simulation Images; Bottom Row is Real Images. To the human eye, the simulated and real scenes are visually highly similar. Still, machine learning systems struggle.}
    \label{fig:dataset}
\end{figure}

\begin{table*}[]
    \centering
    \caption{Overview of Pre-trained Vision Encoders included in the study}
    \begin{tabularx}{\textwidth}{c X c c c c c}
    \toprule
    \textbf{No.} & \textbf{Model Name} & \textbf{Architecture} & \textbf{Embedding Dim} & \textbf{Parameters (M)} & \textbf{Training Type} & \textbf{Pre-Training Data} \\
    \midrule
    1 & CLIP-Base-16 & Transformer & 512 & 51 & Self-supervised & General \\
    2 & CLIP-Base-32 & Transformer & 512 & 151 & Self-supervised & General \\
    3 & CLIP-Large-14 & Transformer & 768 & 432 & Self-supervised & General \\
    4 & DinoV2-B & Transformer & 768 & 86 & Self-supervised & General \\
    5 & EfficientNet B0 & CNN & 1280 & 5.3 & Supervised & General \\
    6 & HRP-ResNet18 & CNN & 512 & 11.7 & Supervised & Manipulation \\
    7 & HRP-ViT & Transformer & 768 & 24 & Supervised & Manipulation \\
    8 & MCR & CNN & 2048 & 5.9 & Self-supervised & Robot Manipulation \\
    9 & MobileNetV3 & CNN & 1280 & 5.4 & Supervised & General \\
    10 & MVP & Transformer & 768 & 43 & Self-supervised & Manipulation \\
    11 & R3M ResNet18 & CNN & 512 & 11.7 & Self-supervised & Manipulation \\
    12 & R3M ResNet34 & CNN & 512 & 21.3 & Self-supervised & Manipulation \\
    13 & R3M ResNet50 & CNN & 2048 & 25.6 & Self-supervised & Manipulation \\
    14 & ResNet18 & CNN & 512 & 11.7 & Supervised & General (ImageNet) \\
    15 & ResNet34 & CNN & 512 & 21.3 & Supervised & General (ImageNet) \\
    16 & ResNet50 & CNN & 2048 & 25.6 & Supervised & General (ImageNet) \\
    17 & ResNet101 & CNN & 2048 & 44.6 & Supervised & General (ImageNet) \\
    18 & Swin Transformer & Transformer & 1024 & 87 & Self-supervised & General \\
    19 & VC1-B & CNN & 768 & 22.6 & Self-supervised & General \\
    20 & VGG-16 & CNN & 4096 & 138 & Supervised & General \\
    21 & VGG-19 & CNN & 4096 & 143 & Supervised & General \\
    22 & VIP & CNN & 1024 & 22.6 & Self-supervised & Manipulation \\
    23 & Vision Transformer (ViT) & Transformer & 768 & 86 & Supervised & General (ImageNet) \\
    \bottomrule
    \end{tabularx}
    \label{tab:encoder_comparison}
\end{table*}

\subsection{The Encoders}
A vision encoder is a function  $f_{\theta}: R^{H \times W \times C} \xrightarrow{} d_{z}$ that maps an image $ x_{i} \in R^{H \times W \times C}$ with height and width being 240 by 320 in our case, and the standard three channels for RGB. The output $f_{\theta}(x_i)$ is a 1-dimensional vector $z_i \in R^{d_x}$ which is supposed to be a low-dimensional, information-dense representation of the contents of the image.

Twenty-three encoders are analyzed across all metrics to evaluate the performance of various model architectures in the Sim2Real gap, making this study contain, to our knowledge, the largest number of models examined in any study within this field \cite{r3m_paper, resnet_paper,efficientnet, vgg_paper, VC1_paper, mobilenet_paper, mcr_paper, ViT_paper, MVP_paper, swin_paper, CLIP_paper, dinov2_paper, VIP_paper,  HRP_paper}. Our encoder selection provides a broad and representative sampling of contemporary visual representation learning approaches, spanning multiple neural network structures, training methodologies, and datasets useful for training more robust models. Each encoder is categorized into Convolutional Neural Networks (CNNs) \cite{CNNs_Intro} and Vision Transformers (ViTs) \cite{ViTs_Intro}. CNNs are traditional approaches to image analysis designed to learn and extract features from images hierarchically. Alternatively, ViTs treat inputs like puzzles, relating pieces to create a holistic view of the image. We also compare models pre-trained on different data sets, some using manipulation datasets like Ego4D, and others using general data sets like ImageNet \cite{5206848}, to see if this pre-training data impacts Sim2Real performance. A complete list of the encoders is provided in Table \ref{tab:encoder_comparison}, including architectural type, training approach, and size.


\subsection{Generating Embeddings (Passing Images with Neural Networks)}
We use each encoder to generate and store embeddings for all sim and real-world images. Embeddings are vector representations of the observed images the encoders generate, abstractly demonstrating learned representations and extracted features from the image. Each pre-trained vision encoder will process images differently due to fundamental differences in the neural network designs, processing methods, and learned weights, demonstrating varying performance in our thorough qualitative and quantitative procedures. All model embeddings are normalized to ensure analyses are on the same scale across all benchmarks.


\subsection{Domain Invariance Score (DIS) Using Centroid Distances}
With our framework, we quantify each encoder's domain invariance using a centroid distance metric, which measures the average difference between generated embeddings of simulated and real images to compare the extent of the Sim2Real gap across models. 

Let $f_\theta:\mathbb{R}^{H\times W\times C}\!\to\!\mathbb{R}^{d}$ be an encoder and $z=f_\theta(x)$ its embedding. To compare encoders with differing output dimensions on a common space, we concatenate all embeddings (sim+real) into $E\in\mathbb{R}^{n\times d}$, standardize features, and project to a fixed dimension $d^\star$ via PCA: $\tilde{E}=\mathrm{PCA}_{d^\star}(\mathrm{Std}(E))$. We then apply per-feature min–max normalization to obtain $\hat{E}$ with components in $[0,1]$. Denote simulated and real subsets by $\hat{E}_{\mathrm{sim}}$ and $\hat{E}_{\mathrm{real}}$, and their centroids
\[
\mu_{\mathrm{sim}}=\frac{1}{n_s}\sum_{j=1}^{n_s}\hat{z}_j^{\mathrm{sim}},\qquad
\mu_{\mathrm{real}}=\frac{1}{n_r}\sum_{i=1}^{n_r}\hat{z}_i^{\mathrm{real}}.
\]
The \emph{Domain Invariance Score (DIS)} is the inverse, dimension-normalized Euclidean gap between centroids,
\begin{equation}\label{eq:dis}
    \mathrm{DIS} \;=\; \Big[\,1-\frac{\lVert \mu_{\mathrm{real}}-\mu_{\mathrm{sim}}\rVert_2}{\sqrt{d^\star}}\,\Big]_+,
\end{equation}
with $[t]_+=\max\{0,t\}$. This choice follows two-sample testing intuition: with a linear kernel, domain discrepancy reduces to the distance between mean embeddings; thus a smaller $\mathrm{DIS}$ on \ref{fig:big_plot} indicates stronger sim–real alignment.

\begin{algorithm}
\caption{Domain Invariance Score (DIS) with PCA + Min–Max + Euclidean}
\label{alg:dis}
\KwIn{Embeddings $E\in\mathbb{R}^{n\times d}$, flags $D\in\{\mathrm{sim},\mathrm{real}\}^n$, target dim $d^\star$}
\BlankLine
$\tilde{E}\leftarrow \mathrm{PCA}_{d^\star}(\mathrm{Std}(E))$ \tcp*{standardize then project}
\For{$j=1$ \KwTo $d^\star$}{ 
  $m_j \leftarrow \min_i \tilde{E}_{ij}$,\quad $M_j \leftarrow \max_i \tilde{E}_{ij}$,\quad
  $\hat{E}_{ij} \leftarrow \frac{\tilde{E}_{ij}-m_j}{\max(M_j-m_j,\; \varepsilon)}$
}
Partition $\hat{E}$ into $\hat{E}_{\mathrm{sim}}$ and $\hat{E}_{\mathrm{real}}$ using $D$\;
$\mu_{\mathrm{sim}} \leftarrow \frac{1}{|\hat{E}_{\mathrm{sim}}|}\sum_{z\in \hat{E}_{\mathrm{sim}}} z$\;
$\mu_{\mathrm{real}} \leftarrow \frac{1}{|\hat{E}_{\mathrm{real}}|}\sum_{z\in \hat{E}_{\mathrm{real}}} z$\;
$d \leftarrow \lVert \mu_{\mathrm{real}}-\mu_{\mathrm{sim}}\rVert_2 \big/ \sqrt{d^\star}$\;
\Return $\mathrm{DIS} \leftarrow \max(0,\, 1 - d)$\;

\end{algorithm}

\subsection{Action Score (AS) Using Probing}
Our framework can quantitatively analyze each encoder's ability to infer actions with a technique we call ``action probing,'' based on the interpretation technique of linear probing \cite{linearprobe}. 

Given frozen encoder embeddings $z_i \in \mathbb{R}^{d_z}$ and ground-truth robot actions $a_i \in \mathbb{R}^{d_a}$, we assess task-relevant information via a linear probe $g_\phi(z)=Wz+b$ trained with mean-squared error on a train split and evaluated on a held-out validation split. The probe solves
\[
\min_{\phi}\; \frac{1}{n_{\mathrm{tr}}}\sum_{i\in \mathcal{D}_{\mathrm{tr}}} \lVert a_i - g_\phi(z_i)\rVert_2^2,
\]
and we define the \emph{Action Score (AS)} from the validation error as
\begin{equation}\label{eq:as}
\mathrm{AS} \;=\; \Big[\,1 - \frac{1}{n_{\mathrm{val}}\,d_a}\sum_{i\in \mathcal{D}_{\mathrm{val}}}\lVert a_i - g_\phi(z_i)\rVert_2^2 \,\Big]_+,
\end{equation}
where $[t]_+=\max\{0,t\}$. Thus, higher $\mathrm{AS}$ on \ref{fig:big_plot} indicates that actions are linearly recoverable from the encoder’s representation (greater task informativeness).

\begin{algorithm}
\caption{Action Score (AS) via Linear Probing}
\label{alg:as}
\KwIn{Train/val embeddings $\{(z_i,a_i)\}$ with $z_i\!\in\!\mathbb{R}^{d_z}$, $a_i\!\in\!\mathbb{R}^{d_a}$; epochs $T$, batch size $B$}
\BlankLine
Initialize linear probe $g_\phi(z)=Wz+b$\;
\For{$t=1$ \KwTo $T$}{
  \For{minibatch $(Z,A)\subset \mathcal{D}_{\mathrm{tr}}$ of size $B$}{
    $\phi \leftarrow \phi - \eta \nabla_\phi \frac{1}{B}\sum \lVert A - g_\phi(Z)\rVert_2^2$
  }
}
Compute validation MSE: $\mathrm{MSE}_{\mathrm{val}} \leftarrow \frac{1}{n_{\mathrm{val}}\,d_a}\sum_{i\in \mathcal{D}_{\mathrm{val}}}\lVert a_i - g_\phi(z_i)\rVert_2^2$\;
\Return $\mathrm{AS} \leftarrow \max(0,\, 1 - \mathrm{MSE}_{\mathrm{val}})$\;
\end{algorithm}
\FloatBarrier


\subsection{Qualitatively Understanding Encoders with GradCAM Saliency Maps}
Grad-CAM \cite{gradcam} is used to visualize encoder attention on robot scenes. We do not present this as a contribution, but as a supporting check: encoders with low DIS should attend similarly to task-relevant regions across sim and real images, while encoders with high AS should focus sharply on manipulators and objects. This ensures our quantitative metrics correspond to meaningful behaviors.

\section{Experimental Results}
\subsection{Analyzing Domain Invariance - Action Inference Plot}
We evaluated all 23 encoders with the proposed metrics, Domain Invariance Score (DIS) and Action Score (AS), and present the joint results in Fig. \ref{fig:big_plot}. Each point corresponds to a model, with the $x$-axis representing DIS and the $y$-axis representing AS. Encoders closer to the top-left of the plot therefore achieve both strong domain invariance and high action informativeness. Marker color distinguishes CNNs (blue) from transformers (orange), marker shape indicates pre-training type (circle for manipulation-specific data, square for general data), and marker size is proportional to parameter count.

Several clear patterns emerge. First, pre-training data exerts a strong influence: encoders trained on manipulation-relevant datasets consistently attain higher AS values, while encoders trained on generic datasets show wider variance. This trend is especially apparent when comparing ResNet baselines to their R3M counterparts, which are trained with manipulation-focused self-supervision—R3M models consistently achieve higher AS, and in most cases, higher DIS as well. Second, parameter count did not correlate with improved performance: larger encoders did not consistently outperform smaller ones in either metric, suggesting that task relevance of pre-training outweighs raw model capacity. Finally, encoder type plays a role: CNN-based models generally achieve higher DIS than ViTs, implying that convolutional structures preserve greater robustness to the sim–real domain shift. 

According to our framework, the highest-performing encoder is MCR (encoder 8) \cite{mcr_paper}, a CNN pre-trained directly on robotic manipulation data, which places near the top-left corner of the plot. This position reflects both strong alignment of sim and real embeddings (high DIS) and effective linear action decoding (high AS), in contrast to encoders such as DinoV2-B that score low on both axes. Together, the joint analysis of DIS and AS offers a quantitative, interpretable benchmark of Sim2Real encoder quality.

\subsection{Comparing Quantitative Results To Qualitative Observations}

\begin{figure}[!htbp]
    \centering
    \includegraphics[width=\linewidth, height=5cm]{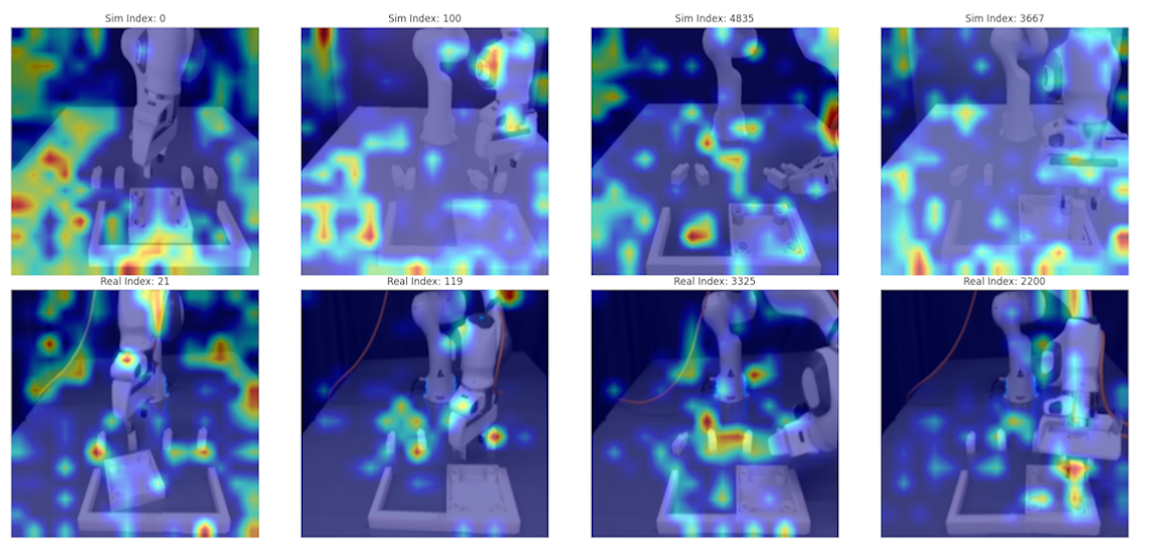}
    \caption{Grad-CAM Maps Generated With DinoV2-B}
    \label{fig:dinov2}
\end{figure}

\begin{figure}[!htbp]
    \centering
    \includegraphics[width=\linewidth, height=5cm]{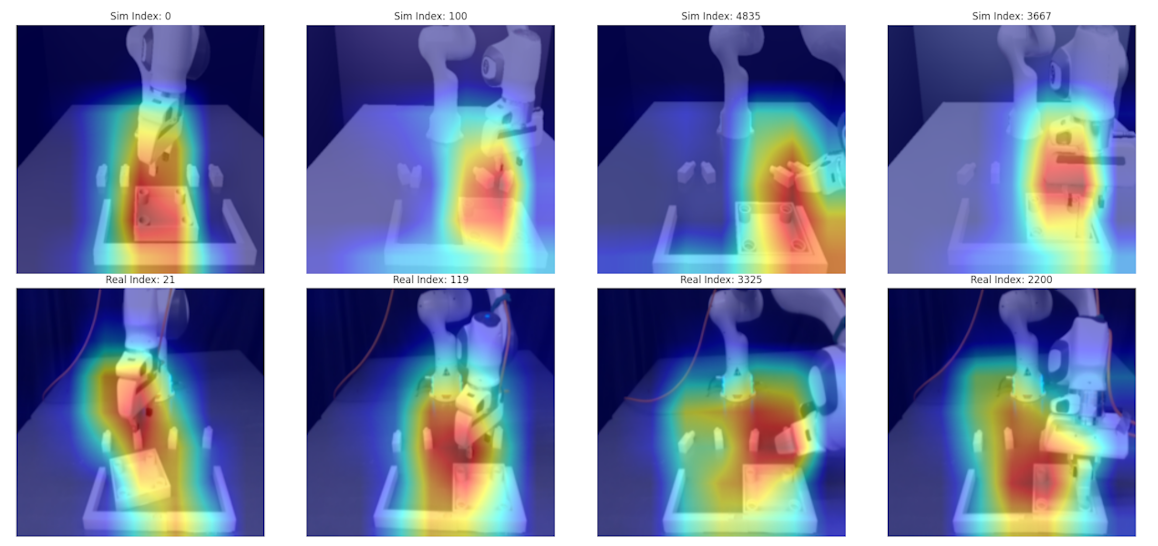}
    \caption{Grad-CAM Maps Generated With MCR}
    \label{fig:mcr}
\end{figure}

We analyze the framework's qualitative results using Grad-CAM saliency maps to contextualize the performance of various encoders on the plot. For each encoder, we generated a grid of saliency maps to observe how they relate to the encoder's quantitative domain invariance and action inference performance. We highlight Fig. \ref{fig:dinov2} and Fig. \ref{fig:mcr}, which contrast the overall worst-performing model (DinoV2-B) and the overall best-performing model (MCR).

\begin{figure*}[!htbp]
    \centering
    \vspace{-10pt}
    \includegraphics[width=\textwidth]{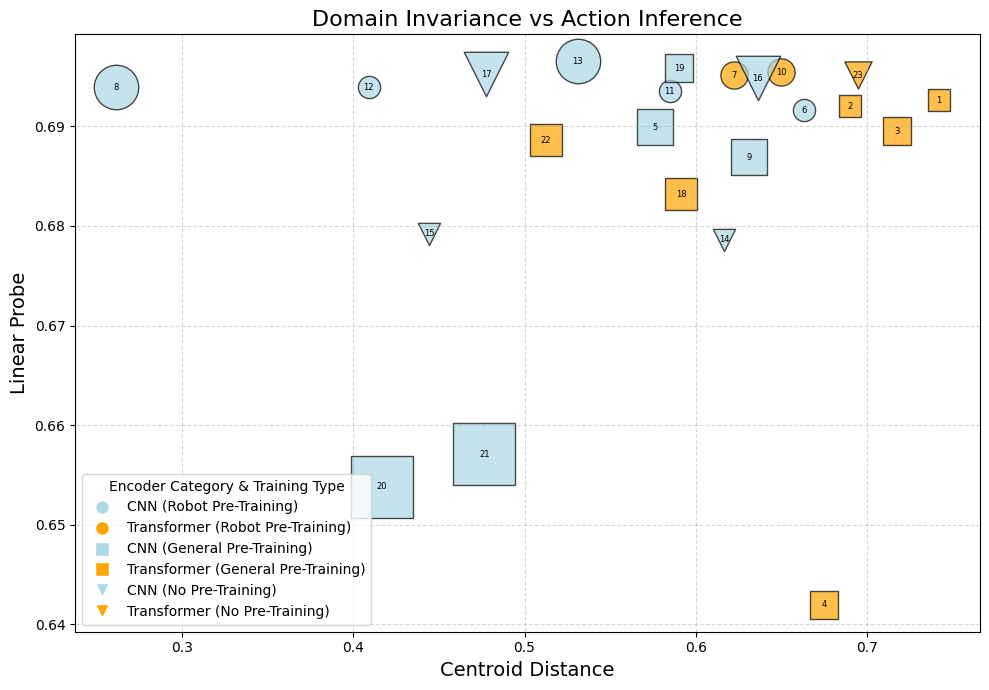}
    \caption{Plot Of Each Encoder's Performance In Domain Invariance and Action Inference, where point size represents the number of parameters, point color represents CNN/ViT-based, and point shape represents manipulation/general pre-training. See Table \ref{tab:encoder_comparison} for corresponding encoder numbers.}
    \label{fig:big_plot}
\end{figure*}

Fig. \ref{fig:dinov2} shows that DinoV2-B (4 in Fig. \ref{fig:big_plot}) does not have a clear focus across any of the maps, indicating low action inference performance, which corresponds with its Action Score. 

Fig. \ref{fig:mcr} shows that MCR (8 in Fig. \ref{fig:big_plot}) has a sharp focus on the robot end-effectors and the objects in the Grad-CAM maps for both simulated and real images. This finding aligns well with its high performance in both quantitative metrics.

\section{Discussion} 
\subsection{Limitations}
While our framework provides a rigorous offline evaluation of encoder transferability, it does not include direct validation on physical robots. As a result, the predictive value of DIS and AS, though theoretically grounded, should ultimately be confirmed in real-world policy deployments. This limitation reflects a broader challenge in Sim2Real research: large-scale robot experiments are costly and difficult to standardize. In addition, although we employed Grad-CAM to provide qualitative interpretability, it is used only as supporting evidence rather than as a primary contribution. Variation in Grad-CAM maps across encoders does not affect the validity of our quantitative metrics, but highlights the inherent noisiness of visual explanation techniques.

\subsection{Future Directions}
Future work can extend this framework in several directions. First, the DIS and AS metrics could form the basis of an open-access benchmark or online database cataloging encoder performance across tasks and domains. Such a resource would allow the robotics community to rapidly compare architectures, training strategies, and pre-training datasets, accelerating progress toward robust visual encoders. Second, these metrics could be incorporated into a reusable software library, enabling practitioners to compute a standardized ``Sim2Real score'' for any encoder by combining DIS and AS into a single cost function. Finally, integrating offline scores with downstream policy transfer studies on real robots would further validate and calibrate the predictive power of our framework, bridging the gap between mathematical analysis and embodied deployment.


\section*{Acknowledgments}
We sincerely thank the Massachusetts Institute of Technology Improbable AI Lab for assisting with this extensive study of pre-trained encoders. Their computational resources and sizable datasets made this possible. We also acknowledge Mr. Lars Ankile for his dedicated mentoring during this investigation. His knowledge and support were crucial in determining the course and methodology of our study. Our effort would not have been possible without their persistent support and generous contributions.




\bibliographystyle{IEEEtran}
\nocite{*}
\bibliography{IEEEabrv,references}{}

\end{document}